\pdfoutput=1

\documentclass[11pt]{article}

\usepackage[]{EMNLP2023}

\usepackage{times}
\usepackage{latexsym}

\usepackage[T1]{fontenc}

\usepackage[utf8]{inputenc}

\usepackage{microtype}

\usepackage{inconsolata}

\usepackage{booktabs}
\usepackage{tabularx}
\usepackage{array}
\usepackage{siunitx}
\newcolumntype{L}{>{\raggedright\arraybackslash}X} 
\usepackage{graphicx}
\usepackage{tcolorbox}
\tcbuselibrary{breakable}

\title{A French Version of the OLDI Seed Corpus}

\author{Malik Marmonier \quad Benoît Sagot \quad Rachel Bawden \\
         Inria, Paris, France\\ \{\texttt{firstname.lastname\}@inria.fr}}

\begin{document}
\maketitle
\begin{abstract}
We present the first French partition of the OLDI Seed Corpus, our submission to the WMT 2025 Open Language Data Initiative (OLDI) shared task. We detail its creation process, which involved using multiple machine translation systems and a custom-built interface for post-editing by qualified native speakers. We also highlight the unique translation challenges presented by the source data, which combines highly technical, encyclopedic terminology with the stylistic irregularities characteristic of user-generated content taken from Wikipedia. This French corpus is not an end in itself, but is intended as a crucial pivot resource to facilitate the collection of parallel corpora for the under-resourced regional languages of France.
\end{abstract}

\section{Introduction}
While state-of-the-art machine translation (MT) has made significant strides, progress has largely been concentrated on a handful of high-resource languages \citep{haddow-etal-2022-survey}. For many of the world's languages, development is hindered by a lack of reliable training data \citep{joshi-etal-2020-state}. Techniques like backtranslation \citep{bertoldi-federico-2009-domain, bojar-tamchyna-2011-improving, sennrich-etal-2016-improving} can help bridge this gap, but they typically require an initial high-quality parallel corpus to ``kickstart'' the process. To address this, the OLDI Seed Corpus was created (originally as the NLLB Seed Dataset) to provide a small but high-quality, professionally translated dataset for dozens of low-resource languages \citep{nllbteam2022languageleftbehindscaling, maillard-etal-2023-small}. The source material consists of approximately 6,000 English sentences sampled from a curated list of core Wikipedia articles, ensuring broad topic coverage \citep{nllbteam2022languageleftbehindscaling}.

The WMT 2025 Open Language Data Initiative (OLDI) shared task builds directly on this effort, inviting the research community to expand these foundational open-source datasets to more languages. Our work answers this call by adding a French partition to the OLDI Seed Corpus. While French is a high-resource language, its selection as a pivot is strategic for our ultimate goal: facilitating the creation of parallel corpora for under-resourced regional languages of France (e.g., Francoprovençal, Occitan, Picard). Practically, translators for these languages are overwhelmingly more likely to have a native command of French than of English. Most importantly, many of these languages exist in a diglossic relationship with French, which serves as the dominant language for most technical and formal domains. A French source text would, therefore, greatly simplify the complex terminological work inherent in translating the encyclopedic content found in the OLDI Seed Corpus, enabling the creation of direct calques and other word-formation strategies that are linguistically and culturally more congruent than those derived from English. With that in mind, we went to great lengths to rigorously verify the French technical terminology throughout the data creation process. By providing this carefully curated French corpus, we aim to establish a solid foundation for future translation efforts.

In addition to the final French partition, we also contribute a supplementary dataset containing the full set of translation hypotheses generated by the nine different MT systems and prompting techniques used in our workflow. This resource, which pairs multiple machine-generated outputs with a final, human-post-edited reference for each source segment, may be of particular use for research on preference optimization and quality estimation in the Wikipedia domain.\footnote{We make this resource publicly available under CC BY-SA 4.0 license: \url{https://github.com/mmarmonier/ACReFOSC}}

\section{Linguistic Overview}
French is a Romance language that evolved from the Vulgar Latin spoken by the inhabitants of northern Gaul after the Roman conquest. Its development was significantly shaped by the Germanic invasions of the 5th century, particularly by the Franks, whose linguistic habits profoundly influenced the phonology and vocabulary of the emerging Gallo-Romance vernacular \citep{rickard_history_2014}. The earliest extant text, the Strasbourg Oaths, dates to 842. Over the subsequent centuries, the dialect of the Île-de-France region, known as Francien, gradually gained prestige due to the political and cultural centrality of Paris. This Parisian variety formed the basis of a standardized literary and administrative language that was progressively imposed throughout the kingdom, through a process legally enforced by the Ordinance of Villers-Cotterêts in 1539. This history of internal linguistic unification through political centralization later provided a model for its imposition as the language of administration and education throughout France's colonial empire.

Today, with over 321 million speakers, French is the fifth most spoken language in the world \citep{OIF2022-LangueFrancaise}. However, this numerical strength, driven almost entirely by African demography, masks a more complex reality. Recent independent assessments point to a rebalancing, with French losing ground to English in several symbolic and high-value domains. In parts of West Africa, its constitutional standing has weakened, with countries like Mali and Burkina Faso downgrading it from an ``official'' to a ``working'' language in 2023--2024. In North Africa, educational policies increasingly favor English, while in major Anglophone countries like the United Kingdom and the United States, learner numbers have seen long-term declines \citep{CollenDuff2025-LTEngland, LusinEtAl2023-MLAEnrollments2021}. Furthermore, English overwhelmingly dominates high-prestige domains such as scientific publishing even within France's own research output \citep{ost_2024_scientific_position_france_en}. Despite these trends, the variety used in this corpus, standard French as spoken in France, remains the most widely recognized norm for formal written communication.

\section{Data Collection}
The shared task guidelines for Seed data contributions permit the use of post-edited machine translation (MTPE). Given the high quality of modern MT systems for the English-French language pair, we adopted this workflow. A full professional translation from scratch was deemed prohibitively expensive. Counterintuitively perhaps for a ``seed'' corpus, the source text represents a non-trivial amount of written content; at 136,656 total words, its length is comparable to that of a novel, falling between Jane Austen's \textit{Pride and Prejudice} and Charles Dickens' \textit{A Tale of Two Cities}. We estimated that a professional translation would have required a substantial budget—around €50,000, not accounting for the added complexity of the varied and often highly technical terminology found in the source segments—that we felt would be more strategically allocated to future translation efforts into the low-resource regional languages of France. The MTPE approach therefore allowed us to produce a highly adequate French corpus while preserving resources for our long-term goals.

\subsection{Machine translation}
The source text was the English partition of the OLDI Seed Corpus. To generate a diverse set of initial translation hypotheses for post-editing, we made use of nine different MT systems and/or prompting techniques. Among these were four ``traditional'' sequence-to-sequence Transformer models \citep{NIPS2017_3f5ee243}. We used a standard bilingual OPUS-MT model \citep{tiedemann-thottingal-2020-opus} trained on a large collection of open parallel corpora. We also used three larger multilingual models: two from the NLLB family, namely the 3.3B-parameter model and the 600M distilled version \citep{nllbteam2022languageleftbehindscaling}, and the 3B-parameter model from the MADLAD-400 project \citep{kudugunta2023madlad400multilingualdocumentlevellarge}. For all four of these systems, translations were generated at the sentence level using beam search with a beam size of 4.

The remaining five translation hypotheses were generated using Large Language Models (LLMs). 

Four hypotheses were generated using Llama 4 Scout, a recently released 109B-parameter Mixture-of-Experts model \citep{meta_ai_llama_2025} with a remarkable 10M-token context window. The first and most straightforward approach was to translate each segment individually, though in a context-informed way. For this, we designed a detailed prompt that instructed the model to act as an expert translator of encyclopedic documents and provided it with a set of guidelines adapted from the official OLDI translation instructions.\footnote{\url{https://oldi.org/translation-guidelines.pdf}} A crucial aspect of the OLDI Seed corpus is that, while its segments are sourced from a limited number of Wikipedia articles, they are not necessarily contiguous. To account for this, the prompt supplied the model with preceding text from the source article as context (up to five segments), while explicitly stating that this context might not be directly adjacent to the segment being translated. We also prompted the model to use a chain-of-thought process before producing the final translation, which was to be enclosed in specific XML tags (<translation>...</translation>) for automatic retrieval (see Appendix~\ref{sec:appendixA1}).

To take advantage of the model's large context window, the other three hypotheses from Llama 4 Scout were generated by translating at the document level. We reconstructed the source documents by grouping all segments from the OLDI Seed corpus that shared the same source URL in their metadata, ordering them by their numerical ID. This full-document text was then translated in a single prompt under three different conditions, with the model explicitly instructed to produce only the final translation without any preceding chain-of-thought. The first setting was identical to the segment-level approach, including the full set of translation guidelines (see Appendix~\ref{sec:appendixA2}). The second was a contrastive ablation where we removed the OLDI guidelines from the prompt, in order to test the model's ability to follow a detailed translator's brief. For the third setting, we again used the full set of instructions, but also provided the model with the complete text of the corresponding French Wikipedia article as additional, in-domain context to inform its translation (see Appendix~\ref{sec:appendixA3}).

The last translation hypothesis was produced by DeepSeek-R1 \citep{deepseekai2025deepseekr1incentivizingreasoningcapability} via its web interface, at the document level, with prompts identical to the first document-level setting used with Llama 4 (guidelines, no corresponding French Wikipedia page; see Appendix~\ref{sec:appendixA2}). Due to the model's safety filters, it refused to translate a handful of documents pertaining to religion, racism, or politics.

A final processing step was required for all document-level translations. The single block of translated text produced by the LLMs had to be segmented and re-aligned with the original source segments. While this process was largely automated, manual intervention was required for approximately 5\% of the documents to correct errors typically resulting from skipped or hallucinated segments, or from the addition of extraneous newlines.

\begin{figure}[h!] 
  \centering
  \includegraphics[width=\linewidth]{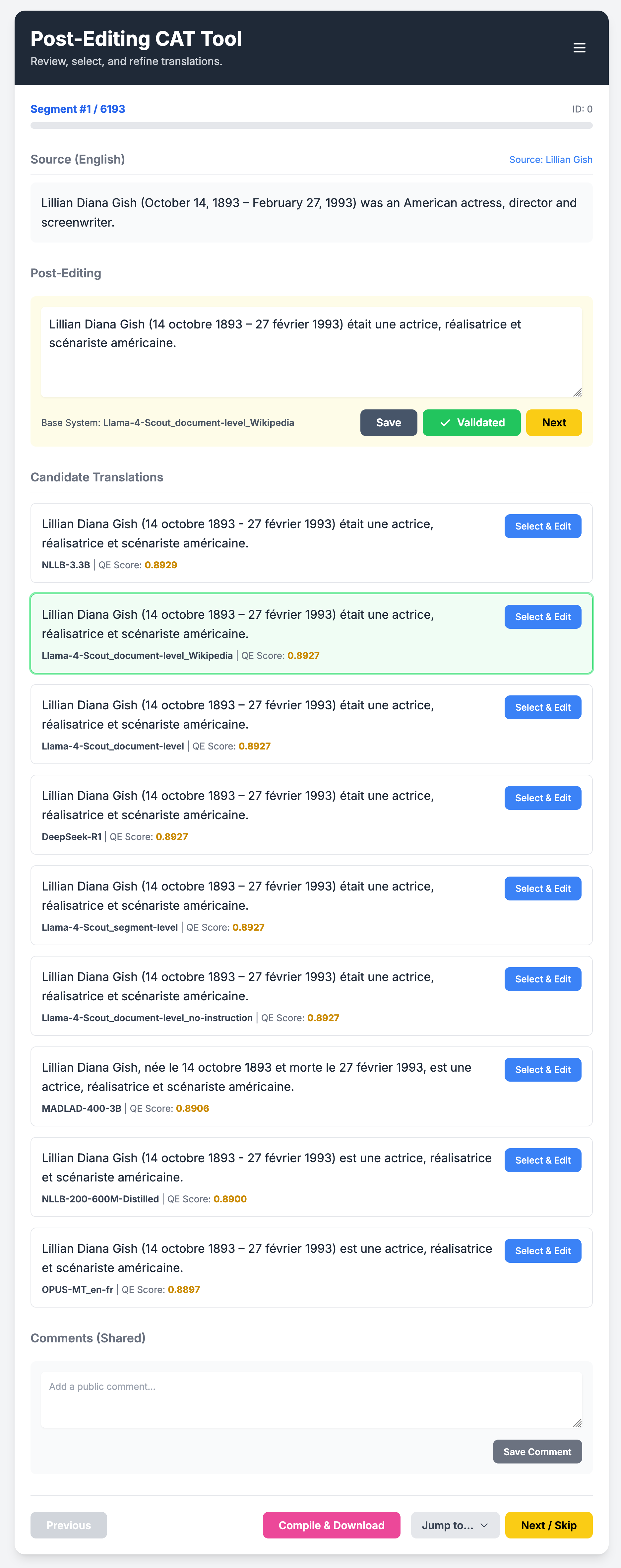}
  \caption{Our custom post-editing interface.}
  \label{fig:fig1}
\end{figure}

\subsection{Human post-edition}
The core of our contribution lies in a meticulous post-editing process, which was performed by two native French speakers with C2-level proficiency in English. A native British English speaker with C2 proficiency in French was also available for consultation to resolve ambiguities in the source text.

To facilitate this work, we developed a custom post-editing interface\footnote{The interface was developed using Vue.js and styled with Tailwind CSS.} (see Figure~\ref{fig:fig1}). For each source segment, the interface presents the user with all nine machine-generated hypotheses, sorted in descending order of their COMET\footnote{wmt22-cometkiwi-da} quality estimation (QE) scores \citep{rei-etal-2022-cometkiwi}. The post-editor can then select the most promising candidate, which populates a text area for refinement. This process allowed for an efficient workflow focused on two primary goals:

\textbf{Fluency.} Improving the naturalness and readability of the French text. A significant challenge throughout the post-editing process was dealing with various types of errors and disfluencies present in the English source text. These issues, likely stemming from the nature of user-generated content and the automated extraction process used to create the corpus, required careful interpretation and, at times, consultation with other language versions of the Seed corpus to resolve ambiguities. The problems ranged from simple typographical errors and ungrammatical constructions to more complex issues like garden-path sentences and segmentation errors that rendered segments nonsensical. Table~\ref{tab:postediting-challenges} provides several examples of these challenges and our resulting post-edited translations.

\begin{table*}[htbp]
\centering
\small
\setlength{\tabcolsep}{6pt}
\renewcommand{\arraystretch}{1.2}
\begin{tabularx}{\textwidth}{@{}l >{\raggedright\arraybackslash}X@{}}
\toprule
\textbf{ID} & \textbf{English Source \& Issue Description} \\
\midrule

2129 &
\textbf{Source:} ``Carleton University and the University of Western Ontario, 1945 and 1946 prospectively, created Journalism specific programs or schools.'' \par
\textbf{Issue:} Typographical, lexical and grammatical errors (``prospectively'' for ``respectively'', missing preposition, capitalization). \par
\textbf{Post-edited:} ``\textit{L'Université Carleton et l'Université Western Ontario ont créé des programmes ou des écoles spécifiques de journalisme, respectivement en 1945 et 1946.}'' \\

\addlinespace

2244 &
\textbf{Source:} ``Without social capital in the area of education, teachers and parents who play a responsibility in a students learning, the significant impacts on their child's academic learning can rely on these factors.'' \par
\textbf{Issue:} Ungrammatical and confusing sentence structure (garden path, unclear pronoun reference). Resolved in part by soliciting the opinion of a native English speaker, and by following the interpretation of the Spanish version of the corpus. \par
\textbf{Post-edited:} ``\textit{Sans capital social dans le domaine de l'éducation, et sans des enseignants et des parents qui jouent un rôle essentiel dans l'apprentissage de l'élève, les impacts significatifs sur l'apprentissage scolaire de leur enfant peuvent dépendre de ces facteurs.}'' \\

\addlinespace

3881 &
\textbf{Source:} ``The first genetically modified ornamentals commercialized altered color.'' \par
\textbf{Issue:} Garden-path sentence structure. Resolved by following the interpretation of the Italian version of the corpus. (We note that the Spanish version of the OLDI Seed has a different analysis of the segment.) \par
\textbf{Post-edited:} ``\textit{Les premières plantes ornementales génétiquement modifiées commercialisées changeaient de couleur.}'' \\

\addlinespace

4326 &
\textbf{Source:} ``When interest rates are very low, the number 0 is included if the interest rate is less than 1\%, e.g.\ ``\% Treasury Stock'', not ``\% Treasury Stock''.)'' \par
\textbf{Issue:} Apparent error in the source text, likely from faulty extraction of a mathematical or financial example. The error was preserved as per the guidelines. \par
\textbf{Post-edited:} ``\textit{Lorsque les taux d'intérêt sont très bas, le nombre 0 est inclus si le taux d'intérêt est inférieur à 1~\%, par ex.\ « \% Treasury Stock », et non « \% Treasury Stock ».)}'' \\

\addlinespace

4539 &
\textbf{Source:} ``Another early globe, the Hunt--Lenox Globe, ca.'' \par
\textbf{Issue:} Sentence segmentation error; the segment is incomplete. \par
\textbf{Post-edited:} ``\textit{Un autre des premiers globes, le globe Hunt--Lenox, env.}'' \\

\bottomrule
\end{tabularx}
\caption{Examples of challenges encountered in the source text during post-editing.}
\label{tab:postediting-challenges}
\end{table*}

\textbf{Accuracy.} Rigorously verifying and correcting the translation of technical terminology. Given the encyclopedic nature of the content, this required systematic external research. The corpus covers a dizzying array of technical topics, from cartography and bionanotechnology to Gothic architecture and Hilbert's problems, and verifying the correct terminology for each was a significant undertaking. Official resources like FranceTerme\footnote{\url{https://www.culture.fr/franceterme}} were often of limited use. We therefore relied on extensive documentary research to find correct or acceptable equivalents in French, a crucial step to ensure the corpus can effectively serve as a pivot for the complex neology that will likely be required for translation into the regional languages of France. This task was made more difficult by the increasing prevalence of low-quality, machine-generated content on the web, or ``AI slop,'' which degrades its utility as a reliable concordancer for the translator. In a handful of cases, we consulted with domain experts to resolve particularly challenging terminological issues. Table~\ref{tab:terminology-segments} provides examples of such challenging segments.

\begin{table*}[htbp]
\centering
\small
\setlength{\tabcolsep}{6pt}
\renewcommand{\arraystretch}{1.2}
\begin{tabularx}{\textwidth}{@{}l >{\raggedright\arraybackslash}X@{}}
\toprule
\textbf{ID} & \textbf{English Source \& Terminological Field} \\
\midrule

4044 &
\textbf{Source (Nanotechnology):} ``Another group of nanotechnological techniques include those used for fabrication of nanotubes and nanowires, those used in semiconductor fabrication such as deep ultraviolet lithography, electron beam lithography, focused ion beam machining, nanoimprint lithography, atomic layer deposition, and molecular vapor deposition, and further including molecular self-assembly techniques such as those employing di-block copolymers.'' \par
\textbf{Post-edited:} ``\textit{Un autre groupe de techniques nanotechnologiques comprend celles utilisées pour la fabrication de nanotubes et de nanofils, celles utilisées dans la fabrication de semi-conducteurs telles que la lithographie ultraviolette profonde, la lithographie par faisceau d'électrons, l'usinage par faisceau d'ions focalisés, la lithographie par nano-impression, le dépôt en couches atomiques et le dépôt moléculaire en phase vapeur, et incluant en outre des techniques d'auto-assemblage moléculaire telles que celles employant des copolymères à diblocs.}'' \\

\addlinespace

4845 &
\textbf{Source (Gothic Architecture):} ``Lancet windows were supplanted by multiple lights separated by geometrical bar-tracery.'' \par
\textbf{Post-edited:} ``\textit{Les fenêtres en lancette furent supplantées par des baies multiples séparées par des remplages géométriques.}'' \\

\bottomrule
\end{tabularx}
\caption{Examples of segments requiring specialized terminological research.}
\label{tab:terminology-segments}
\end{table*}

\subsection{Human validation}
\begin{figure}[h] 
  \centering
  \includegraphics[width=\linewidth]{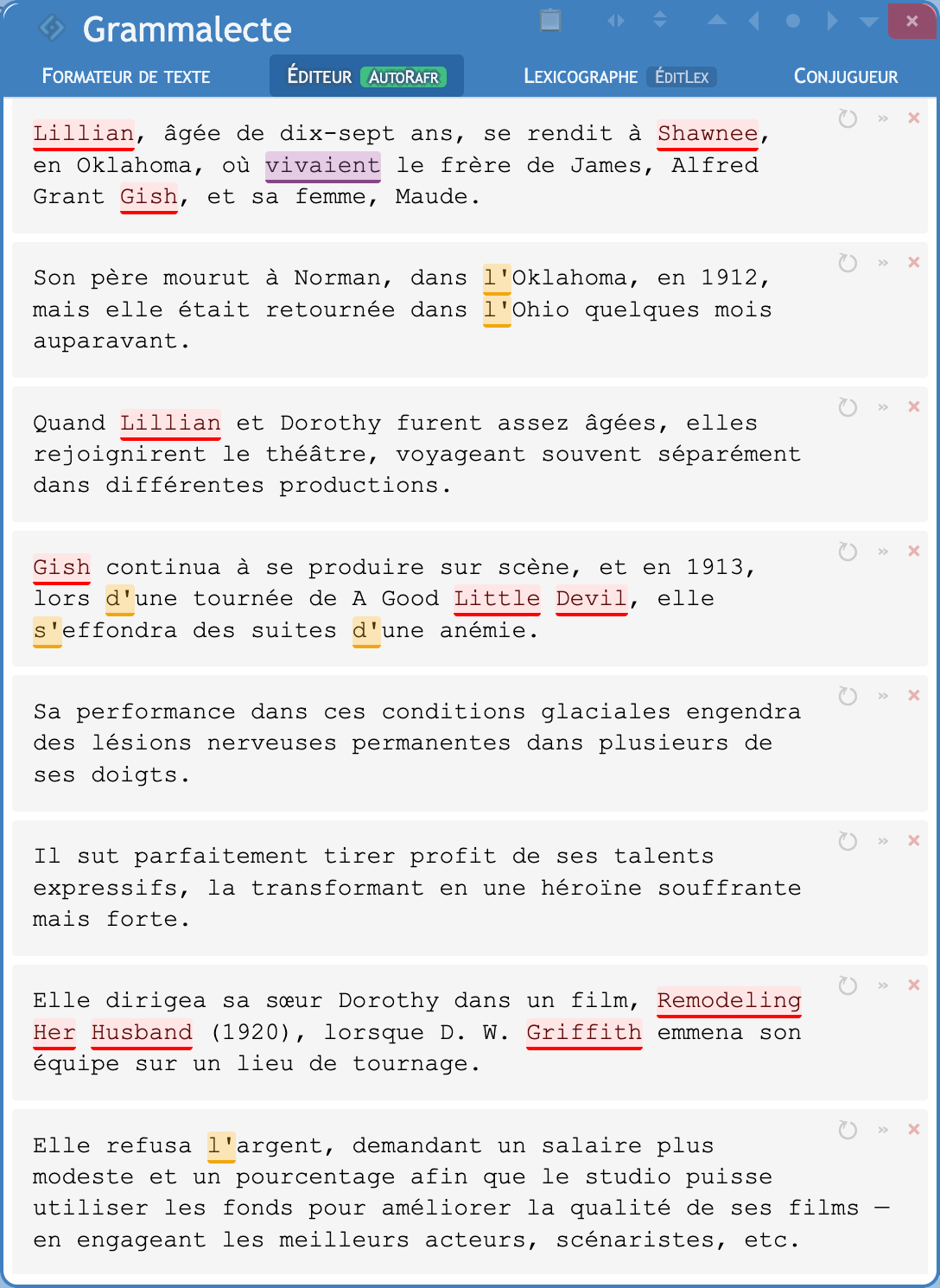}
  \caption{Grammalecte interface used for validation.}
  \label{fig:fig2}
\end{figure}

After post-editing, all segments were processed through the Grammalecte\footnote{\url{https://grammalecte.net/}} grammar checker interface (see Figure~\ref{fig:fig2}) for a final validation pass, correcting any residual spelling or grammatical errors.

As per the shared task guidelines for Seed data, we have ensured that the terms of service for all MT models used allow for the reuse of their outputs. The final dataset is released under the same CC BY-SA 4.0 license as the source corpus.

\section{Discussion}
While we were initially tempted to follow the validation approach of \citet{cols-2024-spanish} by training separate NMT models on our final corpus and on each set of hypotheses, we ultimately judged that for a well-resourced language pair like English-French, and given that the initial MT hypotheses were already of a high standard—failing mostly on specific terminological choices and disfluencies inherited from the source—quality estimation using a state-of-the-art metric would provide a more telling as well as a more environmentally responsible validation of our data.

Although we had used a COMET-based QE model in our post-editing interface, we had observed certain limitations, such as insensitivity to terminological accuracy and a tendency to reward superficial trivial features (like hyphen or apostrophe type) at the expense of more essential features in the candidate translations. For our final validation, we therefore chose MetricX-24\footnote{metricx-24-hybrid-xl-v2p6} \citep{juraska-etal-2024-metricx}, a top-performing hybrid reference-based and reference-free metric. MetricX-24 is trained on human judgements (MQM and DA ratings; cf. \citealt{burchardt-2013-multidimensional, graham-etal-2013-continuous}) and predicts an error score, where lower scores indicate higher quality. This allows for a direct comparison of our final, human post-edited translations against the raw machine-generated hypotheses, thereby quantifying the value added by our human-in-the-loop process.

We scored our final post-edited translations and the raw outputs of the nine systems used to generate initial hypotheses. The results, including 95\% confidence intervals and statistical significance groupings, are presented in Table~\ref{tab:metricx24}.

\begin{table*}[h]
\centering
\small
\begin{tabularx}{0.8\linewidth}{L S[table-format=1.4] l c}
\toprule
\textbf{System} & {\textbf{Avg. Error} $\downarrow$} & \textbf{95\% C.I.} & \textbf{Group} \\
\midrule
Human post-edition                            & 2.0790 & {[2.04, 2.12]} & A \\
\midrule
NLLB-3.3B                                              & 2.2223           & {[2.19, 2.26]} & B \\
MADLAD-400-3B                                          & 2.2290           & {[2.19, 2.27]} & B \\
Llama-4-Scout\_segment-level                           & 2.2437           & {[2.21, 2.28]} & B \\
\midrule
Llama-4-Scout\_document-level\_no-instruction          & 2.3096           & {[2.27, 2.35]} & C \\
Llama-4-Scout\_document-level                          & 2.3198           & {[2.28, 2.36]} & C \\
\midrule
Llama-4-Scout\_document-level\_Wikipedia               & 2.4322           & {[2.38, 2.48]} & D \\
\midrule
NLLB-200-600M-Distilled                                & 2.5332           & {[2.49, 2.58]} & E \\
DeepSeek-R1                                            & 2.5411           & {[2.48, 2.60]} & E \\
\midrule
OPUS-MT\_en-fr                                         & 2.7019           & {[2.65, 2.75]} & F \\
\bottomrule
\end{tabularx}
\caption{MetricX-24 evaluation on the full dataset. Lower is better. Systems in the same lettered group are not statistically significantly different.}
\label{tab:metricx24}
\end{table*}

As noted in Section 3.1, DeepSeek-R1 refused to translate a small number of documents due to its safety filters. To ensure a fair comparison of translation quality on the segments all systems were able to process, we also performed an analysis excluding these 165 segments. The results of this filtered evaluation are presented in Table~\ref{tab:metricx24-excl-refused}.

\begin{table*}[h]
\centering
\small
\begin{tabularx}{0.8\linewidth}{L S[table-format=1.4] l c}
\toprule
\textbf{System} & {\textbf{Avg. Error} $\downarrow$} & \textbf{95\% C.I.} & \textbf{Group} \\
\midrule
Human post-edition                   & 2.0871 & {[2.05, 2.12]} & A \\
\midrule
DeepSeek-R1                                   & 2.2238           & {[2.18, 2.26]} & B \\
NLLB-3.3B                                     & 2.2313           & {[2.19, 2.27]} & B \\
MADLAD-400-3B                                 & 2.2386           & {[2.20, 2.28]} & B \\
Llama-4-Scout\_segment-level                  & 2.2532           & {[2.22, 2.29]} & B \\
\midrule
Llama-4-Scout\_document-level\_no-instruction & 2.3186           & {[2.28, 2.36]} & C \\
Llama-4-Scout\_document-level                 & 2.3302           & {[2.29, 2.37]} & C \\
\midrule
Llama-4-Scout\_document-level\_Wikipedia      & 2.4456           & {[2.39, 2.50]} & D \\
NLLB-200-600M-Distilled                       & 2.5451           & {[2.50, 2.59]} & D \\
\midrule
OPUS-MT\_en-fr                                & 2.7099           & {[2.66, 2.76]} & E \\
\bottomrule
\end{tabularx}
\caption{MetricX-24 evaluation excluding segments refused by DeepSeek-R1. Lower is better. Systems in the same lettered group are not statistically significantly different.}
\label{tab:metricx24-excl-refused}
\end{table*}

Both analyses clearly validate the quality of our contributed dataset. The human post-edited text achieves the lowest average error score by a significant margin, placing it in a statistical group of its own (Group A), confirming the soundness and effectiveness of our manual post-editing and validation process. Among the machine-generated hypotheses, the results on the full dataset (Table~\ref{tab:metricx24}) show a top tier (Group B) consisting of the larger sequence-to-sequence models and the segment-level Llama 4 Scout prompt. However, when excluding the segments DeepSeek-R1 refused to translate (Table~\ref{tab:metricx24-excl-refused}), its relative performance improves significantly, moving it into this top tier of MT systems. This was not surprising, as we were consistently impressed during post-editing by DeepSeek-R1's knowledge of even the most arcane terminology, which we systematically verified through arduous external research ourselves. The remaining rankings are largely consistent across both analyses. Interestingly, all document-level prompting strategies for Llama 4 Scout were slightly less effective than the segment-level approach. The setting that ablated the specific OLDI guidelines (``no-instruction'') performed on par with the standard document-level prompt from the point of view of MetricX-24; a direct comparison reveals that the specific OLDI guidelines had a limited impact on model generation, as these two settings produced identical translations in over 76\% of cases, and for the minority of instances where they differed, the average Translation Edit Rate (TER; cf.~\citealt{olive2005global, snover-etal-2006-study}) was a relatively low 9.48, indicating the variations were typically minor. Counter-intuitively, providing the model with the corresponding French Wikipedia article as additional context resulted in a statistically significant degradation in quality. Finally, the weakest-performing systems across all settings were the smaller distilled NLLB model and the bilingual OPUS-MT model.

Our post-editing logs offer another lens through which to evaluate the raw quality of the MT hypotheses. Of the 6,193 segments in the corpus, 3,043 (49.14\%) had at least one machine-generated hypothesis that was deemed perfect by the post-editors and required no changes. A breakdown by system reveals that DeepSeek-R1 was by far the most reliable, providing the ``perfect'' translation in 2,503 instances, or 40.42\% of all segments in the corpus. The other systems lagged considerably behind, with the various Llama 4 Scout prompting strategies and the larger NLLB and MADLAD models providing the perfect match in 7-9\% of cases, while the smaller NLLB and OPUS-MT models did so less than 5\% of the time.

\section{Related Work}

While data-driven approaches to machine translation have precedents as early as the 1950s \citep{EdmundsonHays1958-ResearchMT}, they did not truly come into their own until the late 1980s and early 1990s with the pioneering work on statistical machine translation (SMT) at IBM (\mbox{\citealt{brown-etal-1990-statistical}}; \mbox{\citealt{berger-etal-1994-candide}}). This early research was heavily reliant on the availability of large parallel corpora, and the French-English language pair played a central role, largely thanks to the Canadian Hansard, a bilingual record of parliamentary proceedings. So influential was this corpus that it established a lasting convention in SMT literature, where the letters $f$ and $e$ became the standard variables to denote source ($f$rench) and target ($e$nglish) language strings in equations, regardless of the actual languages involved.
Our work builds on the modern legacy of these corpus-based approaches. Specifically, we contribute to the OLDI Seed Corpus \citep{nllbteam2022languageleftbehindscaling, maillard-etal-2023-small}, a resource designed to bootstrap translation capabilities for low-resource languages. The OLDI shared task has spurred several recent efforts to expand this dataset, including the creation of Spanish \citep{cols-2024-spanish}, Italian \citep{ferrante-2024-high, zenamt24}, and Bangla \citep{ahmed-etal-2024-bangla} partitions, each contributing to this growing ecosystem of open, multiparallel data.

\section{Conclusion}
In this paper, we have presented our contribution to the WMT 2025 OLDI shared task: a high-quality, human-post-edited French partition of the OLDI Seed Corpus. We have detailed our data creation methodology, which relied on a diverse array of machine translation systems and approaches—from traditional NMT models to various LLM prompting strategies—and a custom-built interface to facilitate an efficient and robust post-editing workflow. Our experimental validation, using the state-of-the-art MetricX-24 QE metric, confirmed that our final, manually-refined corpus is of significantly higher quality than any of the individual machine-generated hypotheses. Our primary motivation for this work is to provide a reliable pivot resource to enable the future development of translation capabilities for the under-resourced regional languages of France. We hope that this dataset, along with the supplementary collection of raw MT outputs, will serve as a valuable contribution to the community and a stepping stone towards this long-term goal.

\section*{Limitations}
A key challenge in this work was navigating the tension between the OLDI translation guidelines, which tend to favor a more literal translation approach, and the need to resolve the stylistic and grammatical disfluencies often present in the source Wikipedia segments. While our post-editing process aimed to produce fluent French, a review of the final data in isolation from the source text reveals that some of these disfluencies, while corrected, may have still carried over into the target segments to some extent. This appears to be a common difficulty when working with this particular source corpus; indeed, a brief review of the recently released Italian \citep{ferrante-2024-high, zenamt24} and Spanish \citep{cols-2024-spanish} partitions suggests that their respective translation teams grappled with similar issues. As this French corpus is intended primarily as a pivot resource, we plan to monitor its use in the translation into regional languages of France and will consider releasing future revisions should any significant issues be surfaced during this process.

\section*{Ethics Statement}

In adherence to the OLDI shared task's commitment to open data, all systems used to generate translation hypotheses were carefully selected to ensure that their terms of service were compatible with the final dataset's release under a CC BY-SA 4.0 license. While we initially considered including hypotheses from popular commercial systems such as Google Translate and DeepL, we ultimately decided against it, as their terms of service prohibit the use of their outputs for the purpose of training other machine translation models.

The generation of translation hypotheses using LLMs is a computationally intensive process with an associated environmental cost. We strove to be mindful of this fact throughout our work.

\section*{Acknowledgements}
This work was funded by the French \textit{Agence Nationale de la Recherche} (ANR) under the project TraLaLaM (``ANR-23-IAS1-0006''), as well as by Inria under the ``\textit{Défi}''-type project COLaF. The last two authors' participation was also partly funded through their chairs in the PRAIRIE institute, supported by the ANR as part of the ``\textit{Investissements d'avenir}'' program under the reference ``ANR-19-P3IA-0001,'' and through the ``\textit{France 2030}'' program under the reference ``ANR-23-IACL-0008.''

This work was also granted access to the HPC resources of IDRIS under the allocation 2025-AD011015117R2 made by GENCI.


\appendix

\section{Prompt Samples}
\label{sec:appendix}
\subsection{Segment-Level Prompt}
\label{sec:appendixA1}
\small
\begin{tcolorbox}[colback=gray!5!white, colframe=gray!75!black, breakable]

You are an expert English-French translator of encyclopedic documents. In translating, you adhere to the following guidelines:

1. Refer to the source document context when available. Context helps clarify meaning, resolve ambiguities, and maintain tone and accuracy in translation.

2. Do not convert any units of measurement. Translate them exactly as noted in the source content.

3. Encyclopedic documents should be translated using a formal tone.

4. Provide fluent translations without deviating excessively from the structure of the source segment.

5. Do not expand or replace information compared to what is present in the source segment. Do not add any explanatory or parenthetical information, definitions, etc.

6. Do not ignore any meaningful text that was present in the source segment.

7. If a named entity in the source language has a canonical equivalent in the target language, use this canonical equivalent.

8. If a named entity in the source language does not have a canonical equivalent in the target language, you may use the source term in your translation.

You are to translate the following English source segment into French:

"Her father died in Norman, Oklahoma, in 1912, but she had returned to Ohio a few months before this."

Here is the segment in some of its original context; please note that the context may include parts of the source document that are not directly adjacent to the main segment, and omissions may not be explicitly marked with ellipses. Nonetheless, this context remains valuable for clarifying meaning, resolving ambiguity, and ensuring consistency in tone and terminology:

Gish was a prominent film star from 1912 into the 1920s, being particularly associated with the films of director D. W. Griffith.

She also did considerable television work from the early 1950s into the 1980s, and closed her career playing opposite Bette Davis in the 1987 film The Whales of August.

The first several generations of Gishes were Dunkard ministers.

Their mother opened the Majestic Candy Kitchen, and the girls helped sell popcorn and candy to patrons of the old Majestic Theater, located next door.

The seventeen-year-old Lillian traveled to Shawnee, Oklahoma, where James's brother Alfred Grant Gish and his wife, Maude, lived.

Her father died in Norman, Oklahoma, in 1912, but she had returned to Ohio a few months before this.

A reminder that the English segment you must translate into French is:

"Her father died in Norman, Oklahoma, in 1912, but she had returned to Ohio a few months before this."

You may reflect on the task at hand and explain your chain of thought prior to producing the translation. IMPORTANT: Do write your translation between tags in the following manner: 
<translation>your translation here</translation>.
\end{tcolorbox}
\normalsize

\subsection{Document-Level Prompt}
\label{sec:appendixA2}

\small
\begin{tcolorbox}[colback=gray!5!white, colframe=gray!75!black, breakable]
You are an expert English-French translator of encyclopedic documents. In translating, you adhere to the following guidelines:

1. Refer to the source document context when available. Context helps clarify meaning, resolve ambiguities, and maintain tone and accuracy in translation.

2. Do not convert any units of measurement. Translate them exactly as noted in the source content.

3. Encyclopedic documents should be translated using a formal tone.

4. Provide fluent translations without deviating excessively from the structure of the source segment.

5. Do not expand or replace information compared to what is present in the source segment. Do not add any explanatory or parenthetical information, definitions, etc.

6. Do not ignore any meaningful text that was present in the source segment.

7. If a named entity in the source language has a canonical equivalent in the target language, use this canonical equivalent.

8. If a named entity in the source language does not have a canonical equivalent in the target language, you may use the source term in your translation.

You are to translate the following English document into French. Please note that the following document may include omissions that are not explicitly marked with ellipses. Do not be perturbed by such minor inconsistencies in the source text. These segments were taken from an English Wikipedia page dedicated to 1. IMPORTANT! Please take careful note of the newline characters, as you will need to reproduce them perfectly in your French translation to allow for the automatic alignment of these segments with their English source.

---- BEGINNING OF ENGLISH SOURCE DOCUMENT ----

1 (one, also called unit, and unity) is a number and a numerical digit used to represent that number in numerals.

In conventions of sign where zero is considered neither positive nor negative, 1 is the first and smallest positive integer.

Most if not all properties of 1 can be deduced from this.

It is thus the integer after zero.

It was transmitted to Europe via the Maghreb and Andalusia during the Middle Ages, through scholarly works written in Arabic.

Styles that do not use the long upstroke on digit 1 usually do not use the horizontal stroke through the vertical of the digit 7 either.

By definition, 1 is the magnitude, absolute value, or norm of a unit complex number, unit vector, and a unit matrix (more usually called an identity matrix).

In category theory, 1 is sometimes used to denote the terminal object of a category.

Since the base 1 exponential function (1x) always equals 1, its inverse does not exist (which would be called the logarithm base 1 if it did exist).

Likewise, vectors are often normalized into unit vectors (i.e., vectors of magnitude one), because these often have more desirable properties.

It is also the first and second number in the Fibonacci sequence (0 being the zeroth) and is the first number in many other mathematical sequences.

Nevertheless, abstract algebra can consider the field with one element, which is not a singleton and is not a set at all.

A binary code is a sequence of 1 and 0 that is used in computers for representing any kind of data.

+1 is the electric charge of positrons and protons.

The Neopythagorean philosopher Nicomachus of Gerasa affirmed that one is not a number, but the source of number.

We Are Number One is a 2014 song from the children's TV show LazyTown, which gained popularity as a meme.

In association football (soccer) the number 1 is often given to the goalkeeper.

1 is the lowest number permitted for use by players of the National Hockey League (NHL); the league prohibited the use of 00 and 0 in the late 1990s (the highest number permitted being 98).

---- END OF ENGLISH SOURCE DOCUMENT ----

Only output the translation directly, religiously respecting new lines. Do not add extraneous new lines.
\end{tcolorbox}
\normalsize

\subsection{Document-Level Prompt with Corresponding French Wikipedia Article}
\label{sec:appendixA3}

\small
\begin{tcolorbox}[colback=gray!5!white, colframe=gray!75!black, breakable]

You are an expert English-French translator of encyclopedic documents. In translating, you adhere to the following guidelines:

1. Refer to the source document context when available. Context helps clarify meaning, resolve ambiguities, and maintain tone and accuracy in translation.

2. Do not convert any units of measurement. Translate them exactly as noted in the source content.

3. Encyclopedic documents should be translated using a formal tone.

4. Provide fluent translations without deviating excessively from the structure of the source segment.

5. Do not expand or replace information compared to what is present in the source segment. Do not add any explanatory or parenthetical information, definitions, etc.

6. Do not ignore any meaningful text that was present in the source segment.

7. If a named entity in the source language has a canonical equivalent in the target language, use this canonical equivalent.

8. If a named entity in the source language does not have a canonical equivalent in the target language, you may use the source term in your translation.

The document you will translate consists in segments taken from an English Wikipedia page dedicated to North. Here is what appears to be the corresponding French Wikipedia page (back-matter sections have been removed). It might provide you with the correct terminology and equivalent named entities, pay close attention to these aspects as you read this French text.

---- BEGINNING OF FRENCH WIKIPEDIA ARTICLE ----

Le nord est un point cardinal, opposé au sud.

== Étymologie ==

De l’ancien haut-allemand nord provenant de l’unité linguistique proto-indo-européenne « ner- » qui signifie « gauche », se rapportant sans doute à la gauche du soleil levant.

Le nom de la divinité scandinave Njörd, ayant régné sur une partie du monde pendant un âge d’or, est lié à cette racine[réf. souhaitée]. Cette divinité était connue des Romains sous le nom de Nerthus et avait donné son nom à une des îles du bout du monde, Nérigon.

En latin, Septemtriones signifie les sept bœufs. L'astérisme le plus brillant de l'actuelle constellation de la Grande Ourse, était autrefois une constellation à part entière appelée constellation des sept bœufs. Ce groupement d'étoiles permettait de trouver l’étoile polaire et donc le Nord avec une bonne précision.

Le terme septentrion est un synonyme vieilli de nord, faisant référence à cette constellation qui indiquait la direction du nord aux Romains ; mais l’adjectif septentrional, qui en découle, reste très usité.

== Géographique et magnétique ==

Il existe deux nord. Le premier est magnétique (l’axe de symétrie cylindrique du champ magnétique), le second est géographique (l’axe de rotation de la Terre). Ces deux points ne se trouvent pas au même endroit. Mesuré en 2007 par le projet « Poly-Arctique », le pôle Nord magnétique est situé à 83° 57' 00'' N, 121° 01' 12'' O. Il se trouve à 673 km du pôle Nord géographique et ayant une vitesse moyenne de déplacement de 55 km/an (soit une moyenne d'environ 150 m/jour ou 6 m/h). À l'été 2010, il a été estimé qu'il n'était plus qu'à 550 km du pôle Nord géographique.

La différence d’angle que l’on peut observer sur la boussole entre ces deux nord est appelée déclinaison magnétique. Cette différence varie avec le temps.

Sur les cartes traditionnelles et en particulier les cartes de l’Institut national de l'information géographique et forestière (IGN), les méridiens (lignes noires verticales) pointent le nord géographique (NG) ; il y a donc lieu de tenir compte de la déclinaison magnétique pour s’orienter sur la carte à l’aide d’une boussole (NM). Le croquis situé dans la légende de la carte indique la valeur de la déclinaison pour la carte et pour une année donnée, car le pôle magnétique migre en permanence, réduisant chaque année la valeur de la déclinaison (0,8 degré/an).

Certains cartographes ont contourné cette complication en construisant des cartes tenant compte de cette déclinaison : le nord (N) de la carte ainsi que les lignes verticales en bleu ou en noir pointent le nord magnétique (de la même manière que l’aiguille de la boussole).

La position du nord magnétique a changé plusieurs fois dans l’histoire de la Terre ; la dernière inversion du champ magnétique terrestre s’est produite il y a 780 000 ans.

En l’absence de boussole, le moyen traditionnel pour repérer le nord le soir ou la nuit est de se référer à l’étoile polaire dans l'hémisphère nord ou à la croix du Sud dans l'hémisphère sud. Le jour, il est possible de se référer à la position du Soleil en fonction de l'heure locale. Lorsque le ciel est couvert, observer la mousse ou les vents dominants est peu fiable.

== Typographie ==

Les points cardinaux, qu'ils soient utilisés comme nom ou comme qualificatif, s'écrivent avec :

une majuscule lorsqu'ils font partie d'un toponyme ou désignent une région ;

une minuscule s'ils désignent une direction, une exposition, une orientation.

=== Articles connexes ===

Sud

Point cardinal

---- END OF FRENCH WIKIPEDIA ARTICLE ----

You are to translate the following English document into French. Please note that the following document may include omissions that are not explicitly marked with ellipses. Do not be perturbed by such inconsistencies in the source text. IMPORTANT! Please take careful note of the newline characters, as you will need to reproduce them perfectly in your French translation to allow for the automatic alignment of these segments with their English source.

    ---- BEGINNING OF ENGLISH SOURCE DOCUMENT (this line need not be translated) ----

    North is one of the four compass points or cardinal directions.

Septentrionalis is from septentriones, "the seven plow oxen", a name of Ursa Major.

For example, in Lezgian, kefer can mean both "disbelief" and "north", since to the north of the Muslim Lezgian homeland there are areas formerly inhabited by non-Muslim Caucasian and Turkic peoples.

On any rotating astronomical object, north often denotes the side appearing to rotate counter-clockwise when viewed from afar along the axis of rotation.

But simple generalizations on the subject should be treated as unsound, and as likely to reflect popular misconceptions about terrestrial magnetism.

    ---- END OF ENGLISH SOURCE DOCUMENT (this line need not be translated) ----

IMPORTANT! Only output the translation directly, religiously respecting new lines. Do not add extraneous new lines. Do not skip any segment.

\end{tcolorbox}
\normalsize

\end{document}